# Color Image Segmentation using Adaptive Particle Swarm Optimization and Fuzzy C-means

Narayana Reddy A
*Department of Computer Science and Engineering*
*National Institute of Technology, Mizoram*
Aizawl, India
narayanareddy500@gmail.com

Ranjita Das
*Department of Computer Science and Engineering*
*National Institute of Technology, Mizoram*
Aizawl, India
ranjita.nitm@gmail.com

*Abstract*—Segmentation partitions an image into different regions containing pixels with similar attributes. A standard non-contextual variant of Fuzzy C-means clustering algorithm (FCM), considering its simplicity is generally used in image segmentation. Using FCM has its disadvantages like it is dependent on the initial guess of the number of clusters and highly sensitive to noise. Satisfactory visual segments cannot be obtained using FCM. Particle Swarm Optimization (PSO) belongs to the class of evolutionary algorithms and has good convergence speed and fewer parameters compared to Genetic Algorithms (GAs). An optimized version of PSO can be combined with FCM to act as a proper initializer for the algorithm thereby reducing its sensitivity to initial guess. A hybrid PSO algorithm named Adaptive Particle Swarm Optimization (APSO) which improves in the calculation of various hyper parameters like inertia weight, learning factors over standard PSO, using insights from swarm behaviour, leading to improvement in cluster quality can be used. This paper presents a new image segmentation algorithm called Adaptive Particle Swarm Optimization and Fuzzy C-means Clustering Algorithm (APSOF), which is based on Adaptive Particle Swarm Optimization (APSO) and Fuzzy C-means clustering. Experimental results show that APSOF algorithm has edge over FCM in correctly identifying the optimum cluster centers, there by leading to accurate classification of the image pixels. Hence, APSOF algorithm has superior performance in comparison with classic Particle Swarm Optimization (PSO) and Fuzzy C-means clustering algorithm (FCM) for image segmentation.

## I. Introduction

Image segmentation forms the basis for identifying the objects in the image and forming a contextual relationship between the objects identified. Fuzzy C-means is one of the classic clustering algorithms used in image segmentation to obtain a color quantized version of the source image, which is used in further processing to segment the image. Many inherent complexities arise on applying Fuzzy C-means in the domain of image segmentation. One of the preliminary issues is to pre-specify the number of clusters, which is a daunting task in real life image datasets. Others include FCM falling into local optima due to randomized selection of initial centers. Many studies based on evolutionary algorithms have been proposed to solve this random initialization issue. J Kennedy and Russell [1] proposed an optimization algorithm based on swarm intelligence. Following this, Omran et al. [2] applied this PSO in image classification. In recent years, hybrid versions of PSO which include optimizations in velocity and position calculations shown to have a better performance in comparison with the original PSO [1]. S Gao et al. [3] used the output of PSO to initialize K-means and obtained better results. He also proved that PSO + K-means combination is the better one when compared with K-means + PSO combination. Haiyang Li et al. [4] used the same approach while updating the inertia weight in consecutive iterations of PSO. In this study we aim to use Fuzzy C-means instead of K-means to further improve image segmentation results. Paired with Adaptive PSO, Fuzzy C-means will improve the results as it classifies a data point with a fractional membership value rather than a strict 0 or 1 as in K-means.

## II. Methodology

Fuzzy C-means algorithm introduced by Bezdek et al. [5] groups the data points $x_i, i = 1,2,3, \ldots N$ into $C$ clusters, according to the objective function

$$J_m = \sum_{i=1}^{N} \sum_{j=1}^{C} u_{ij}^m \|x_i - c_j\|^2, \quad (1)$$

where $c_j$ represents the prototype value of the $j^{th}$ cluster, $u_{ij}$ is the degree of membership of $c_j$ in the cluster j, $m$ is any real number greater than 1. In order to minimize the objective function, high membership values are assigned to those pixels, whose intensities are situated close to the prototype values of their clusters. $u_{ij}$ is calculated using (2) and $c_j$ is calculated using (3).

$$u_{ij} = 1 \Big/ \sum_{k=1}^{C} \left(\frac{\|x_j - c_j\|}{\|x_i - c_k\|}\right)^{\frac{2}{m-1}}, \quad (2)$$

$$c_j = \frac{\sum_{i=1}^{N} u_{ij}^m \cdot x_i}{\sum_{i=1}^{N} u_{ij}^m}, \quad (3)$$

This algorithm is converged when the value of $J_m$ stops changing in the subsequent iterations. J Kennedy and Russell [1] introduced PSO in which optimal solution of a problem is abstracted as a particle without mass and volume flying in N-dimensional space. A group of particles is called as Particle Swarm or simply a Swarm. Each particle has its own flight velocity, spatial position and a fitness value. Consider the position of the particle $i$ is $X_i = (x_{i1}, x_{i2}, \ldots, x_{id})$ and velocity is $V_i = (v_{i1}, v_{i2}, \ldots, v_{id})$ in $d$-dimensional space. In an iterative process, the current personal optimal solution *p-best* is $P_i = (p_{i1}, p_{i2}, \ldots, p_{id})$ and the current global optimal solution *g-best* is $P_g = (p_{g1}, p_{g2}, \ldots, p_{gd})$. The velocity and position of a particle can be calculated by using (4) and (5) respectively.

$$v_{ij}(t+1) = wv_{ij}(t) + c_1 r_1 [p_{ij} - x_{ij}(t)] \\ + c_2 r_2 [p_{gj} - x_{ij}(t)] \quad (4)$$

$$x_{ij}(t+1) = x_{ij}(t) + v_{ij}(t+1), \quad j = 1,2,3, \ldots, d \quad (5)$$

where $w$ is the inertia weight. It represents the amount of current flight velocity inherited to the next flight velocity.

$c_1$ represents particle's self-learning capability and $c_2$ represents particle's social learning capability. In the PSO algorithm, $c_1$ and $c_2$ are constants whose values range between 0 and 4. In general, $c_1 = c_2 = 2.0$. $r_1$ and $r_2$ are both random numbers uniformly distributed between 0 and 1. Haiyang Li et al. [4] used the concept of dynamic inertia weight in PSO. In this work, we extend upon this idea to achieve better segmentation results. PSO performance can be improved by dynamically changing the inertia weight. Keeping inertia weight constant may cause algorithm to converge at local optima and linearly decreasing inertia weight may even overshoot from the optimal point. To overcome these drawbacks, the inertia weight can be dynamically calculated using (6).

$$w = \begin{cases} \frac{(w_{max} - w_{min}) \times (f_i - f_{min})}{f_{avg} - f_{min}}, & f_i \geq f_{avg} \\ w_{max}, & f_i < f_{avg} \end{cases} \quad (6)$$

where $w_{max}$ is the maximum inertia weight and $w_{min}$ is the minimum inertia weight. $f$ is current fitness of particle $i$. $f_{avg}$ is current average fitness of the swarm and $f_{min}$ is minimum fitness value of all particles in the swarm. This method has the following advantages:

1. Particles having fitness value less than the average fitness value, are assigned with lower inertia weight, thereby decreasing flight velocity to maintain their position.

2. Particles having fitness value more than the average fitness value, are assigned with higher inertia weight, thereby increasing flight velocity to quickly move closer their fittest neighbours.

The two learning factors of PSO algorithm are generally constants. Inappropriate values assigned to these factors may result in undesired output. To produce better results, we calculate the learning factors $c_1$ and $c_2$ using (7) and (8) respectively.

$$c_1 = c_1^i + \frac{c_1^f - c_1^i}{n_{max}} \times n, \quad (7)$$

$$c_2 = c_2^i + \frac{c_2^f - c_2^i}{n_{max}} \times n, \quad (8)$$

Where $c_1^i > c_2^i$ and $c_1^f < c_2^f$. $n_{max}$ represents the maximum number of iterations, n represents the current iteration. By using the above equations, the values of learning factors become larger or smaller along with the no. of iterations, thus we can treat them as Adaptive learning factors. During the initial stages of optimization, the particles will have particles exhibit greater self-learning capability when compared to the social-learning capability. In later stages of optimization, particles tend to have strong social learning-capability when compared with self-learning capability, accelerating convergence to global optimal solution.

For initializing the swarm (i.e. solution set), the initial cluster centers are randomly selected $M$ pixels out of total $N$ pixels of the image $X$, which act as the solution set of the optimization problem. After the cluster centers are determined, the remaining pixels in image $X$ should be assigned to these $M$ classes according to the following clustering criteria. Let $x_i$ be the $i^{th}$ data point in data point set $X$ and $c_j$ be the $j^{th}$ cluster center. If $\| x_i - c_j \| = min \| x_i - c_k \|, k = 1,2,3 \ldots M$ then assign $x_i$ to the $j^{th}$ cluster. The fitness of a particle is evaluated using the following equation:

$$f_i = \sum_{i=1}^{N} \sum_{j=1}^{M} \|x_i - c_j\|^2, \quad (9)$$

Where $f_i$ represents fitness of particle $i$, $N$ is the total no. of pixels, $M$ is the no. of particles in the solution set, $x_i$ represents the $i^{th}$ pixel, $c_j$ represents the $j^{th}$ cluster center. The average fitness of the swarm $f_{avg}$ is calculated as

$$f_{avg} = \frac{1}{N} \sum_{i=1}^{N} f_i, \quad (10)$$

The degree of convergence in the swarm can be measured by calculating the swarm fitness variance $\delta^2$ as follows:

$$\delta^2 = \frac{1}{N} \sum_{i=1}^{N} (f_i - f_{avg})^2, \quad (11)$$

Gradually, in the optimization process, the particles fitness will tend to be identical. When this happens, the algorithm is said to be converged and the swarm fitness variance $\delta^2$ will be reduced to a certain range which means the algorithm has been close to the global optimal solution. This optimal solution is used as the initial cluster centers of the FCM algorithm. Fig. 1 presents the step by step process of proposed algorithm.

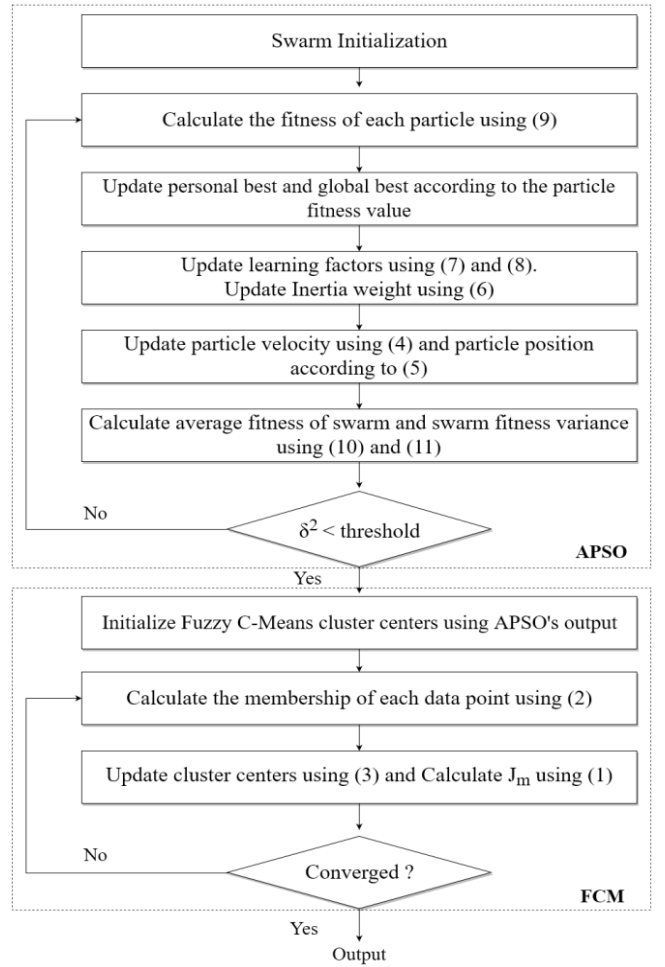

Fig. 1. Process of the proposed APSOF algorithm

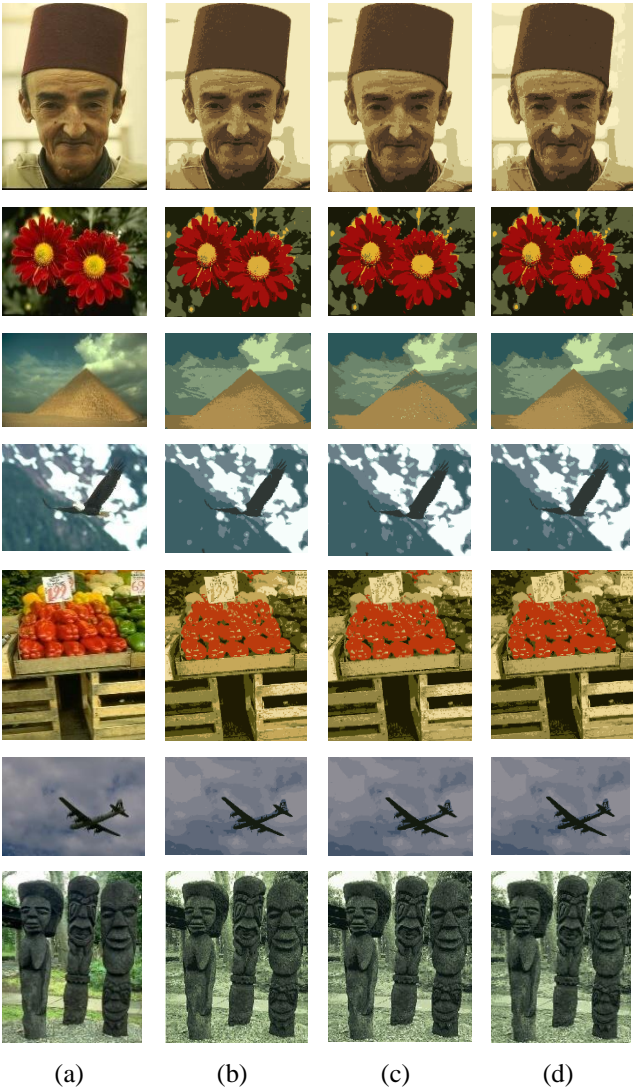

(a)      (b)      (c)      (d)

Fig. 2. Image Segmentation results of three algorithms. (a): original images, (b): K-means, (c): Fuzzy C-means, (d) APSOF.

## III. RESULTS AND DISCUSSION

The APSOF algorithm has been tested using images from The Berkeley Segmentation Dataset and Benchmark [6], which is a standard dataset for the application of image segmentation. Most of the images in this dataset are real life images and are resized to reduce the file size. In this paper seven out of the all the tested images are chosen to be analysed in depth. In order to compare the results, two classic clustering algorithms, namely K-means and Fuzzy C-means are used as comparisons. The initial number of clusters varies depending on the image and generally lies between 5 to 10 at maximum. In the Fuzzy C-means algorithm and APSOF, the value of $m$ is 2. The segmentation results are illustrated in Fig. 2. The performance of APSOF can be analysed from two aspects: qualitative analysis and quantitative analysis. They are presented in the following sections.

### A. Qualitative analysis

The qualitative analysis is basically done to evaluate the image quality through unaided visual inspection, based on the ability to identify interesting regions or objects in an image. From the Fig. 2, we can easily infer that, APSOF has an edge over K-means and Fuzzy C-means in the segmentation results.

TABLE I. NORMALIZED $J_m$ VALUES OF FCM AND APSOF

| Image Name | Fuzzy C-means | APSOF |
|---|---|---|
| Human face | 1.1638 | 0.8362 |
| Flowers | 1.0244 | 0.9756 |
| Pyramid | 1.2016 | 0.7984 |
| Eagle | 1.0011 | 0.9989 |
| Fruits | 1.0951 | 0.9049 |
| Plane | 1.0347 | 0.9653 |
| Zumba | 1.1436 | 0.8564 |

We will analyse image by image to get the visual differences in the segmented images.

From the first row of Fig. 2 (i.e., Human Face), we can see that the portion near eyes and mouth is well differentiated from rest of the face in APSOF when compared with others. Also, the background artefacts, especially on the left side of the face are clearly segmented in APSOF, while it is unclear in other algorithms segmentation result.

Moving to the second row of Fig. 2 (i.e., Flowers), It can be observed that the colors are bright and vibrant in the output of APSOF when compared to Fuzzy C-means, which tends towards the darker hue. K-means, in this case has blown out the background color entirely.

For the Pyramid image as shown in the third row of Fig. 2, APSOF algorithm has the best segmentation results in the region of the pyramid's top and the left end of the clouds. Fuzzy C-means has mis-classified the top portion of the pyramids as a part of the clouds. Although K-means has similar results of APSOF, it lags in the classification of pixels in the corner portions of the pyramids and the clouds.

For the Eagle image as shown in the fourth row of Fig. 2, All the algorithms have produced similar segmentation results except for the blobs near the bottom mid part of the image which can be neglected as they don't have much significance considering the total image.

When we move on to the fifth row of Fig. 2 (i.e., Fruits), the colors of the fruits are brighter in APSOF result. Even the edges are sharper in APSOF when compared with others result. The dark region near the below boxes is correctly shaded with a darker hue in the APSOF result.

Coming to the sixth row of Fig. 2 (i.e., Plane), we can see that the region in the clouds near the head of the plane is clearly segmented in the APSOF when compared to the same portions in the segmentation results of K-means and Fuzzy C-means respectively.

Moving to the last row of Fig. 2 (i.e., Zumba), we can clearly see that the regions on the face and the mid parts of the stone carvings are clearly segmented in APSOF even though they are very noisy in the original image. Even Fuzzy C-means and K-means produced similar results, they tend to misclassify pixels in the face portion.

Overall, based on the qualitative analysis above, it's clear that APSOF algorithm has outperformed K-means and Fuzzy C-means and is superior to them.

### B. Quantitative analysis

The performance of APSOF and Fuzzy C-means can be numerically compared using the final $J_m$ values of both algorithms. For the purpose of easy comparison, the values are normalized by dividing each individual $J_m$ value with the average of both $J_m$ values for each of the output. By the information provided in TABLE I, it's clear that $J_m$ is minimized well in the case of APSOF in most of the cases. For the image of Eagle, there is only slight difference in the $J_m$ values as the segmented outputs of APSOF and Fuzzy C-means are mostly similar.

## IV. CONCLUSION

Fuzzy C-means is a standard algorithm used in image segmentation but sensitive to initialization. Hybrid algorithms like APSOF can improve over the standard Fuzzy C-means to obtain better image segmentation. The performance comparisons indicate that this algorithm is superior to K-means and Fuzzy C-means but with a little sustainable computation cost. The efficiency of APSOF is required to be improved to meet the needs of real-time applications. In future works, spatial information can be incorporated to further improve the segmentation efficiency and strengthen susceptibility of the algorithm to noise in the image data.